\documentclass{article}

    \PassOptionsToPackage{numbers, sort, compress}{natbib}
\usepackage[sort&compress]{natbib}
\usepackage[final]{neurips_2025}

\usepackage{array}
\usepackage{tabularx}

\newcommand{\OurMethod}{LangHOPS}
\newcolumntype{Y}{>{\centering\arraybackslash}X}

\usepackage[utf8]{inputenc} 
\usepackage[T1]{fontenc}    
\usepackage{hyperref}       
\usepackage{url}            
\usepackage{booktabs}       
\usepackage{amsfonts}       
\usepackage{nicefrac}       
\usepackage{microtype}      
\usepackage{xcolor}         
\usepackage{subcaption}
\usepackage{float}
\usepackage{graphicx}
\graphicspath{{figures/}}
\usepackage{booktabs}
\usepackage{comment}
\usepackage{amssymb} 
\usepackage{multirow}
\usepackage{underscore}
\usepackage{pifont}
\usepackage{enumitem}
\usepackage{amsmath} 
\usepackage{array}
\usepackage{tabularx}
\usepackage{wrapfig}
\usepackage{amssymb}
\usepackage{subcaption}
\usepackage{listings}
\usepackage{xcolor}
\usepackage{enumitem}
\newcommand{\crv}[1]{\textcolor{black}{#1}}

\title{\OurMethod{}: Language Grounded Hierarchical Open-Vocabulary Part Segmentation}

\author{%
  Yang Miao \\
  {\footnotesize INSAIT, Sofia University}\\ {"St. Kliment Ohridski"} 
  \And
  Jan-Nico Zaech \\
  {\footnotesize INSAIT, Sofia University}\\ {"St. Kliment Ohridski"} 
  \And
  Xi Wang \\
  {\footnotesize INSAIT, Sofia University}\\ {"St. Kliment Ohridski"}\\{ ETH Zurich, TU Munich}
  \AND 
  Fabien Despinoy \\
  {\footnotesize Toyota Motor Europe}
  \And
  Danda Pani Paudel \\
  {\footnotesize INSAIT, Sofia University}\\ {"St. Kliment Ohridski"} 
  \And
  Luc Van Gool \\
  {\footnotesize INSAIT, Sofia University}\\ {"St. Kliment Ohridski"} 
}

\begin{document}

\maketitle
\begin{figure}[ht]
  \vspace{-5pt}
  \centering
  \includegraphics[width=0.99\linewidth]{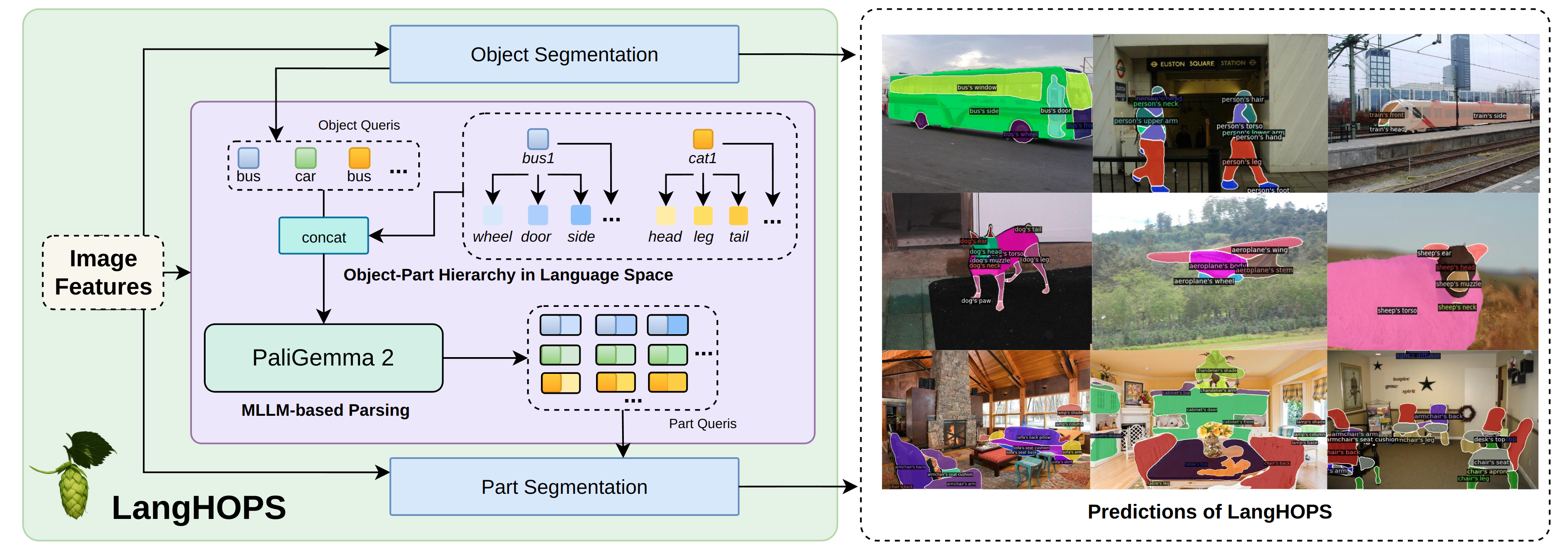}
  \caption{
  Given a 2D image and user queries of candidate object-part categories, our method \OurMethod{} grounds the hierarchy between objects and parts in language space and subsequently leverages a Multimodal Large Language Model to break down the segmented objects into parts.
  }
  \vspace{-3pt}
  \label{fig:teaser}
\end{figure}
\begin{abstract}
We propose \OurMethod{}, the first Multimodal Large Language Model~(MLLM)-based framework for open-vocabulary object–part instance segmentation. 
Given an image, \OurMethod{} can jointly detect and segment hierarchical object and part instances from open-vocabulary candidate categories.
Unlike prior approaches that rely on heuristic or learnable visual grouping, our approach grounds object–part hierarchies in language space. It integrates the MLLM into the object-part parsing pipeline to leverage its rich knowledge and reasoning capabilities, and link multi-granularity concepts within the hierarchies. 
We evaluate \OurMethod{} across multiple challenging scenarios, including in-domain and cross-dataset object-part instance segmentation, and zero-shot semantic segmentation. 
\OurMethod{} achieves state-of-the-art results, surpassing previous methods by $\textbf{5.5\%}$ Average Precision (AP)~(in-domain) and $\textbf{4.8\%}$~(cross-dataset) on the PartImageNet dataset and by $\textbf{2.5\%}$ $mIOU$ on unseen object parts in ADE20K~(zero-shot).
Ablation studies further validate the effectiveness of the language-grounded hierarchy and MLLM-driven part query refinement strategy.
The code will be released \href{https://insait-institute.github.io/langhops.github.io/}{here}. 
\end{abstract}

\section{Introduction}

2D instance segmentation is a well-established computer vision research field and has experienced significant progress in object-level instance segmentation in the past decades~\cite{maskrcnn_2017, cheng2021mask2former, zou2023seem, zhang2025psalm, hyperseg}. 
While recent efforts have expanded toward higher-level reasoning from visual input~\cite{lai2024lisa, hyperseg, OMGLLaVA}, the growing demand for finer semantic understanding has led to increased interest in part-level segmentation~\cite{wei2023ovpart, sun2023vlpart, yuan2024ospreypixel, nguyen2025calico}.
Unlike object-level segmentation, part-level understanding introduces new challenges, as it requires richer contextual awareness, reasoning about object-part relationships, and task-dependent interpretation.
For example, a car can break down into coarse-grained components such as the body and wheels, or further delineated into finer-grained elements, including windows, doors, headlights, mirrors, or screws, depending on the downstream task and desired granularity.
\par
Open-Vocabulary object-Part Instance Segmentation~(OVPIS) emerges as a promising approach to address this challenge and has gained increasing interest in recent years. 
Unlike open-vocabulary object–part semantic segmentation~\cite{wei2023ovpart, choi2024partclipseg, partcatseg}, which assigns part labels to pixels without distinguishing between multiple part instances, object-part instance segmentation requires detecting and segmenting object and part instances separately. 
This introduces additional complexity, as the model must establish part–whole relationships at the instance level and maintain consistent grouping between objects and their constituent parts.
In contrast to closed-vocabulary settings that rely on predefined object-part lists, 
open-vocabulary models aim to generalize to unseen part categories and novel compositions, which is a key capability for real-world generalization.
Among existing works, SAM \cite{sam_2023_ICCV, ravi2024sam2} relies on handcrafted object-part and subpart heuristics for part-level segmentation. However, it does not offer control over the semantic granularity of the parts.
Recent works~\cite{ren2024grounded, zhang2024evfsamearlyvisionlanguagefusion} extend SAM with a text prompt module to guide the segmentation process, but lack modeling of object–part hierarchies, which limits their ability to reason about relationships between objects and corresponding parts.

Moving beyond interactive or prompt-tuned variants of SAM, a separate line of work focuses on open-vocabulary part segmentation by leveraging vision–language models.
OV-Parts~\cite{wei2023ovpart}, PartCLIPSeg~\cite{choi2024partclipseg}, and PartCATSeg~\cite{partcatseg} implement object-part hierarchical reasoning implicitly in CLIP embedding space~\cite{clip} and enable zero-shot transfer to novel part categories. However, the performance of these methods is constrained by the limitations of CLIP in compositional and part-level understanding~\cite{wei2023ovpart, baron2024extendclip_partattribute, concept_association_bias_2023}. 
PartGLEE~\cite{li2024partglee} explicitly models object-part structures using a Q-Former and performs joint object and part instance segmentation. 
Nevertheless, it has suboptimal segmentation performance in open-vocabulary scenarios since the Q-Former module lacks mechanisms to handle part granularity variations.
\crv{Addressing this limitation is essential for improving generalization in real-world applications, where part granularity naturally varies across contexts and user intentions. For example, operating a laptop may require segmenting coarse parts such as the lid, while repairing it demands finer segmentation of detailed components such as screws or hinges.}
\par
In contrast, we propose \OurMethod{}, a novel framework that leverages language-grounded hierarchy and integrates MLLM for the task of OVPIS, as shown in Fig.~\ref{fig:teaser}.
\OurMethod{} embeds object–part hierarchies directly in the language space, producing language-grounded part queries with object context.  
Those queries are further processed by a MLLM to link compositional object-part concepts and to generate adaptive segmentation queries.
To verify the performance of \OurMethod{}, we conduct experiments in multiple settings~(in-domain, cross-dataset and zero-shot) and on multiple dataset~(PartImageNet, PascalPart-116 and ADE20K). 
As a result, \OurMethod{} significantly outperforms baselines by $\textbf{5.5\%}$ AP~(in-domain) and $\textbf{4.8\%}$ AP~(cross-dataset) on the PartImageNet dataset and by $\textbf{2.5\%}$ $mIOU_{\mathrm{unseen}}$ on ADE20K (zero-shot). 
Experiment also shows the advanced scalability of \OurMethod{} with improvement by $\textbf{10.0\%}$ AP on PartImageNet when trained on more dataset, 
Ablation study shows that \OurMethod{} have object-part synergy that part-level instance segmentation can improve object segmentation by \textbf{$\textbf{5.4\%}$} AP.
In summary, our key contributions are:
\begin{itemize}[noitemsep, topsep=0pt, leftmargin=1em]
    \item We propose \OurMethod{}, the first framework integrating an MLLM for the task of OVPIS. 
    \item We propose language-space grounded object-part hierarchy modeling for part query representation and link the multi-granularity concepts with an MLLM to enable context-aware and accurate object-part parsing.
    \item We conduct experiments and demonstrate superior performance of \OurMethod{} in in-domain, cross-dataset, and zero-shot settings, as well as its scalability when on larger datasets. Notably, we show for the first time that part-level supervision can significantly enhance object-level segmentation.
\end{itemize}

\section{Related Work}
\par\noindent\textbf{2D Object-Part Segmentation} aims to jointly detect and segment both objects and their semantic parts, while preserving the hierarchical structure between them~\cite{geus2021partawarepanopticseg, he2021partimagenet, zhou2017ade20k}. 
This task goes beyond traditional object-level understanding~\cite{ding2023maskclip, xu2022groupvit, liang2023open, zhou2025camsam2, fu2024objectrelator, chen2025splitmatch, zhong2025omnisamom, cao2025unlocking} by introducing part-level granularity within object instances.
This topic has gained attention~\cite{partcatseg, li2024partglee, geus2024taskalignedpps, li2024ppsformer, muralidhara2024jppf, wang2023hipie, pan2023ops} due to its potential in downstream applications such as image editing~\cite{kawar2023imagic,ling2021editgan} and robotics~\cite{nair2022r3m, chen2024urdformer}.
TAPPS~\cite{geus2021partawarepanopticseg} extends Mask2Former~\cite{cheng2021mask2former} to predict jointly objects and parts with a set of shared queries. However, it is limited to a fixed set of predefined categories. PartCLIPSeg~\cite{choi2024partclipseg} applies a two-stage strategy for part-level semantic segmentation by first extracting mask proposals and then applying CLIP~\cite{clip} to classify the masked image crops.
\crv{
Nevertheless, CLIP-based approaches such as PartCLIPSeg~\cite{choi2024partclipseg} and OV-Part~\cite{wei2023ovpart} often exhibit suboptimal performance in fine-grained part segmentation, largely due to CLIP’s limited capacity for compositional reasoning and explicit modeling of object–part hierarchies~\cite{wei2023ovpart, baron2024extendclip_partattribute, concept_association_bias_2023}. 
More recently, PartCATSeg~\cite{partcatseg} introduces a cost aggregation framework with a compositional loss and DINO-based structural guidance to enhance part-level image–text alignment and structural understanding. 
However, this method still lacks an explicit, language-grounded mechanism for representing hierarchical object–part relationships, which is essential for robust compositional generalization.}
Separately, PartGLEE~\cite{li2024partglee} adopts a different two-stage pipeline that first segments object instances and then parses object queries into parts using a Q-Former. 
\crv{
Since the Q-Former in PartGLEE~\cite{li2024partglee} is not explicitly aware of part granularity during training or inference, it struggles to adapt across datasets with differing levels of annotation detail. 
For instance, a model trained on fine-grained parts such as “eye,” “nose,” and “ear” for cats in Pascal-Part-116 performs poorly on PartImageNet, where the same category is annotated only with coarser parts (“head,” “body,” “foot,” and “tail”).}
Although \cite{li2024partglee, partcatseg} incorporate object-level context into part segmentation, they do not entirely leverage the hierarchical relationships between objects and parts from candidate category definitions, consequently limiting their overall performance. In contrast, \OurMethod{} explicitly embeds the object-part hierarchies in language space to guide the MLLM for object-part parsing, as detailed in Sec.~\ref{subsec:part_query}.


\par\noindent\textbf{Open-Vocabulary Segmentation} requires models to detect and segment object parts from novel categories guided by free-form text descriptions, without relying on category-specific training data.
Early works, such as MaskCLIP~\cite{ding2023maskclip} and GroupViT~\cite{xu2022groupvit}, initiated this paradigm by using Vision Language Models (VLMs) to transfer knowledge from text supervision to pixel-level tasks. 
Follow-up methods~\cite{liang2023open, zhong2022regionclip} further enhance this capability by introducing text embeddings into mask prediction, contrastive learning, or region-level alignment. These approaches demonstrate the potential of using language as a flexible and scalable supervision signal for segmentation tasks. 
However, most of the existing works~\cite{ding2023maskclip, xu2022groupvit, liang2023open, zhong2022regionclip, pan2023priorPretrieval, pan2023remote_retrieval, ren2023masked, ren2024sharing} focus on object-level semantics, consequently lacking fine-grained part-level reasoning.
OV-Part~\cite{wei2023ovpart} and VLPart~\cite{sun2023vlpart} establish benchmarks for open-vocabulary part segmentation by augmenting existing datasets with part-level annotations~\cite{he2021partimagenet, chen2014pascalpart, zhou2017ade20k, ramanathan2023paco}. Although recent methods~\cite{sun2023vlpart, wei2023ovpart, li2024partglee, partcatseg} make progress towards an open-vocabulary setting, they still exhibit limited generalization, particularly in zero-shot and cross-dataset scenarios where both the label space and data distribution differ from those seen during training.
\OurMethod{} leverages MLLMs and object-part hierarchies to improve generalization and accuracy in the OVPIS task, setting a new benchmark in the open-vocabulary zero-shot and cross-dataset settings.

\par\noindent\textbf{MLLM-based Image Segmentation} integrates multimodal language models into image segmentation tasks, unlocking strong performance in various domains such as open-vocabulary panoptic segmentation, referring segmentation, interactive segmentation, and reasoning-based segmentation~\cite{lai2024lisa, zhang2025psalm, hyperseg, OMGLLaVA}. 
LISA \cite{lai2024lisa} introduces ``reasoning segmentation'', allowing MLLMs to generate the mask token in response to complex and implicit textual queries. 
PSALM~\cite{zhang2025psalm} extends LLMs with a vision encoder and a mask decoder with a flexible input prompt to handle diverse segmentation tasks. 
OMG-LLaVA~\cite{OMGLLaVA} proposes an end-to-end MLLM-based framework capable of image-, object- and pixel-level understanding including pixel-level segmentation.  
While effective, these methods focus on object-level understanding and lack the ability to decompose objects into fine-grained semantic parts.
Osprey~\cite{yuan2024ospreypixel} achieves part-level visual understanding but relies on off-the-shelf class-agnostic part masks~(e.g. from SAM~\cite{sam_2023_ICCV}) and cannot control over part granularity.
More recently, CALICO~\cite{nguyen2025calico} leverages MLLM for multi-image part-focused object comparison by identifying unique and common parts of certain object across images.  
In contrast, \OurMethod{} is the first framework to leverage MLLMs for open-vocabulary object-part instance segmentation, enabling fine-grained parsing at the instance level, beyond the semantic and multi-image settings explored in prior work.




\section{Method}
\begin{figure}
\begin{center}
\includegraphics[width=0.95\linewidth]{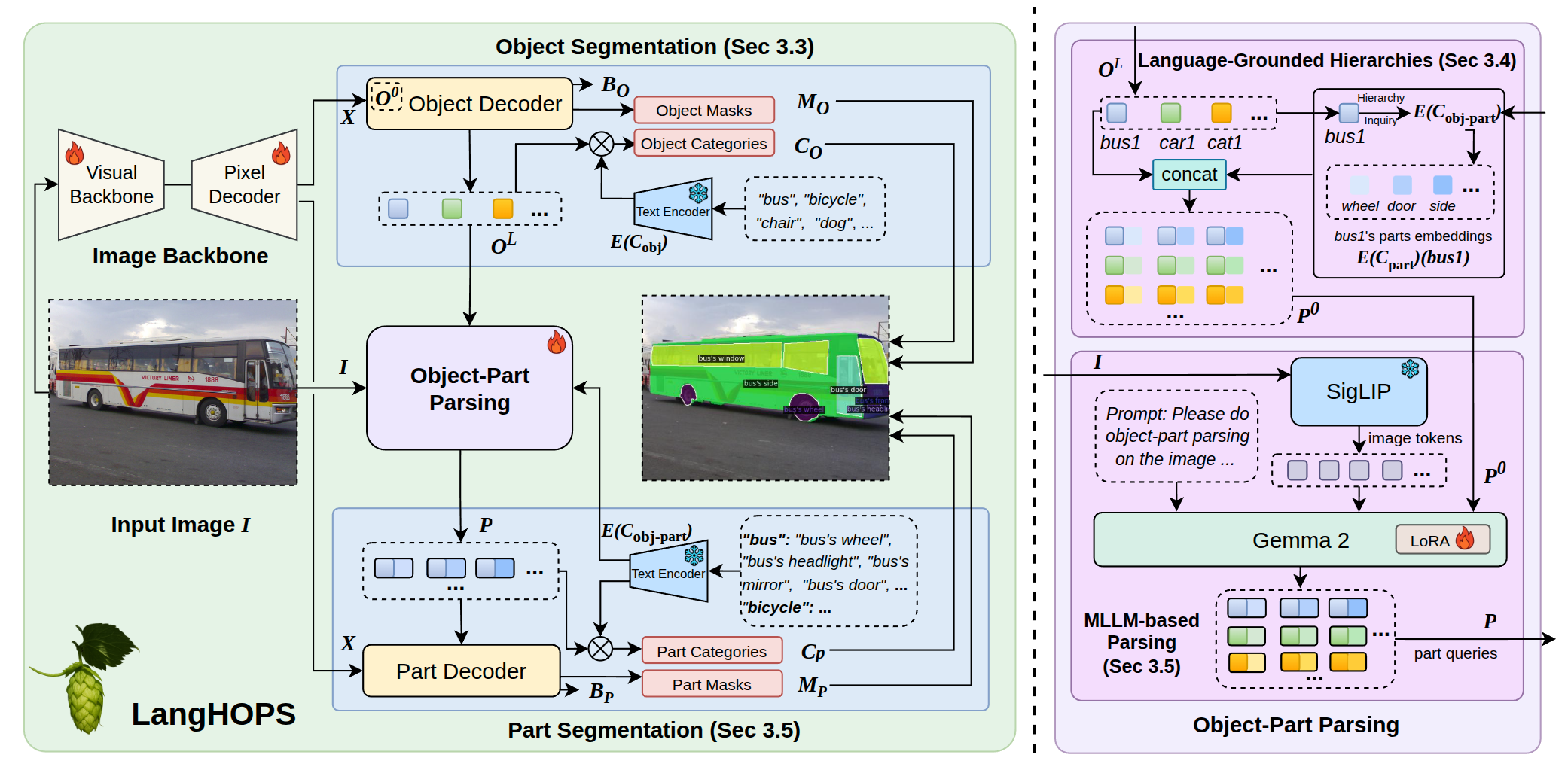}
    \vspace{-5pt}
    \caption{\textbf{\OurMethod{} framework.} The left block illustrates the overall architecture, with an image backbone, an object segmentation module, object-part parser and a part segmentation module. The right block illustrates the ideas on the object-part parser, consisting of a "Language-Grounded Hierarchies" module embedding the object-part hierarchy in language space, and a "MLLM-based Parsing" module producing the part queries for segmentation using a MLLM.}   \label{fig:our_framework}
\end{center}
\vspace{-15pt}
\end{figure}

\subsection{Problem Definition}
OVPIS aims to segment an image into distinct object-level instances and object-specific part-level instances, with the capability to generalize in novel object-part categories. 
Given an image and user-defined open-vocabulary object-part categories, the model outputs masks and categories of objects with their corresponding parts~(e.g., “bus 1”, “bus 1’s headlight 1”, “bus 1’s headlight 2”, etc.). 
Note that, in contrast to the semantic part segmentation task proposed in \cite{wei2023ovpart}, OVPIS also distinguishes between different instances of the same object category. 
For open-vocabulary segmentation, we adopt the commonly used settings in prior work~\cite{sun2023vlpart} where the model takes images and ground-truth mask pairs, for one set of object and part categories
$\mathbf{C}_{\text{}}^{\text{train}}$ during training, and segment objects and parts of novel categories $\mathbf{C}_{\text{}}^{\text{novel}}$ during inference. 

\subsection{Method Overview}
Our model is illustrated in Fig.~\ref{fig:our_framework}.  
Given an input RGB image $\mathbf{I}$ with a set of user-defined open-vocabulary candidate objects and part categories $\mathbf{C}_{\text{}}$, the model outputs the masks and categories of the segmented instances of objects and parts, as well as the object-part hierarchies between the instances. 
Specifically, our framework in Fig.~\ref{fig:our_framework} is composed as follow:
\textbf{Object Segmentation.} 
We derive the initial object queries $\mathbf{O^0}$ with prior information from image features $\mathbf{X}$, and apply the object decoder utilized in ~\cite{li2024partglee} together with CLIP text encoder to obtain predictions of object-level categories $\mathbf{C_{O}}$, bounding boxes $\mathbf{B_{O}}$ and segmentation masks $\mathbf{M_{O}}$. \\
\textbf{Language-grounded Hierarchies.} 
We first extract hierarchies between objects and parts from the input candidate categories $\mathbf{C}_{\text{}}$ and encode them in CLIP’s language space $\mathbf{E(C_{\text{}})}$. 
Subsequently, given predictions of objects' categories, we construct initial part-level queries $\mathbf{P^0}$ by retrieving object-conditioned part embeddings from $\mathbf{E(C_{\text{}})}$, and concatenating them with object queries, enabling context-aware and granularity-adaptive object-part parsing. \\
\textbf{MLLM-based Parsing.} 
We leverage a MLLM to refine initial object-part-concatenated queries $\mathbf{P^0}$ by linking visual concepts with object-part-concatenated queries through structured prompt guidance, producing enriched part queries $\mathbf{P}$ that capture hierarchical part relationships across both language and visual domains for subsequent decoding. \\
\textbf{Part segmentation.}
We use $\mathbf{P}$ and $\mathbf{X}$ as inputs to the part decoder with the same structure as the object decoder and predict categories $\mathbf{C}_{P}$, bounding boxes $\mathbf{B_{P}}$ and masks $\mathbf{M}_{P}$ of part instances.
\par
\crv{
By integrating the language-grounded hierarchies and MLLM into the object-part parsing framework, our object and part segmentation modules are tightly coupled.
Joint information between parts and objects is utilized through the following information flow:
For each segmented object, a set of part queries is constructed by a concatenation of the object embedding 
generated in the object decoder and the language-encoded part description.
These queries are processed by the MLLM, generating queries capable of open-vocabulary part segmentation.
The processed queries are used in the part decoder to segment each requested part.
The details are provided in the following subsections.}

\subsection{Object Segmentation}\label{subsec:obj_seg}
Following PartGLEE~\cite{li2024partglee}, we apply the transformer decoder implemented in \cite{maskdino} as object decoder. 
Object queries $\mathbf{O^0} \in R^{N\times D_q}$ are initialized given priors from the multi-scale image features $\mathbf{X_s} \in R^{D_s\times \frac{H}{2^s}\times\frac{W}{2^s} }, s = \{2, 3, 4, 5\} $, with 
$N$ as hyper-parameter denoting the number of object queries. 
Next, $L$ layers of a deformable transformer decoder module~\cite{maskdino} are applied for cross-attention computation between $\mathbf{X_s}$ and $\mathbf{O^i}$, as well as self-attention of $\mathbf{O^i}$, $i \in [1, L]$.
The output queries $\mathbf{O}^L$ are utilized to perform object-level detection, classification and segmentation with 3 separate prediction head.  
For detection, a 3-layer MLP is utilized to map $\mathbf{O}^L$ to the coordinates of the bounding boxes $\mathbf{B^O} \in R^{N\times4}$:
\begin{equation}
    \mathbf{B^O} = MLP(\mathbf{O}^L).
\end{equation}
For open-vocabulary object-level classification, we apply the CLIP text encoder to process the user-defined candidate object categories and obtain the object-level class embeddings:
\begin{equation}
    \mathbf{E}({\mathrm{\mathbf{C}}_{\mathrm{obj}}}) = CLIP_{\mathrm{text}}(\mathrm{\mathbf{C}}_{\mathrm{obj}})
\end{equation}
Then the classification logits are calculated by:
\begin{equation}
    \mathbf{S^O} = f_{CO}(\mathbf{O}^L) \cdot \mathbf{E}({\mathrm{\mathbf{C}}_{\mathrm{obj}}}),  
\end{equation}
where $f_{CO}$ is the linear layer mapping $\mathbf{O}^L$ to CLIP embedding space.
By taking the maximum logits over the candidate categories, semantic categories predictions, $\mathbf{C^O}$, are obtained.
For object segmentation, masks are generated by calculating the inner-product between $\mathbf{O}^L$ and the dense mask features $f_{M}(\mathbf{X_2})$ obtained with a 2D convolutional network $f_{M}$ on the dense features $\mathbf{X_2}$:
\begin{equation}
    \mathbf{M^O} = f_{MO}(\mathbf{O}^L) \cdot f_{M}(\mathbf{X_2}), 
\end{equation}
where $f_{MO}$ is 3-layer MLP mapping queries into mask features's embedding space. 

\subsection{Language-grounded Hierarchies}\label{subsec:part_query}
Following object segmentation, the next objective is to decompose each segmented object into its corresponding part-level instances.
This requires first modeling the relationship between objects and their constituent parts.
PartGLEE~\cite{li2024partglee} addresses this by introducing a set of learnable, universal parsing queries that, together with object queries, are processed by a Q-Former to generate a fixed number of part queries for each object.
However, such Q-Former-based object-part parsing method has inherent limitations. First, it lacks of context awareness as it does not incorporate user-defined open-vocabulary categories $\mathbf{C}_{\text{}}$ during object-part parsing. 
Consequently, the model may fail to generalize across domains where the definition of parts differs (e.g., coarse vs. fine-grained part sets). 
Second, the Q-Former-based method suffers from limited generalization from data priors. 
In fact, it has to be entirely trained and lacks external knowledge, which makes it highly dependent on the distribution and coverage of the training data. 
%
\par
To effectively address these issues, we explicitly model the hierarchical object-part structure from $\mathbf{C}_{\text{}}$ in the well-generalizable language space. 
Specifically, given one object $\mathbf{O}^L \in R^{1\times{D_q}}$ and its predicted object category ${C_{o}}$ from Sec. \ref{subsec:obj_seg}, 
we query its potential part categories using $\mathbf{C}_{\text{}}$: $\mathrm{\mathbf{C}}_{\mathrm{part}}$. 
For example, if an object with the query $\mathbf{O}^L$ is classified as a "bus", then we retrieve all the parts belonging to "bus"  from $\mathbf{C}_{\text{}}$~("bus's wheel", "bus's window", "bus's door", etc).
Subsequently, we encode the retrieved part categories into the CLIP text embedding space:
\begin{equation}
    \mathbf{E}(\mathrm{\mathbf{C}}_{\mathrm{part}}(C_{o})) = \{ CLIP_{\mathrm{text}}(C^p_o) \}, C^p_o \in \mathrm{\mathbf{C}}_{\mathrm{part}}(C_{o})
\end{equation}
where $\mathrm{\mathbf{C}}_{\mathrm{part}}(C_{o})$ represents part categories belonging to a corresponding object category $C_{o}$.
Ultimately, the embedding of each candidate part is concatenated with the corresponding object query separately as the initial part queries, with both object-level context and part-level language priors:
\begin{equation}
    \mathbf{p_i^0} =  (\ \mathbf{O}^L\  \|  \ f_{CO}(\mathbf{e^i_o} ) \ ), \mathbf{e^i_o} \in \mathbf{E}(\mathrm{\mathbf{C}}_{\mathrm{part}}(C_{o})),
\end{equation}
where $(\cdot \| \cdot)$ represents the concatenation of two tensors. 
\crv{
The query $\mathbf{p_i^0}$ is beyond pure text embeddings.
Instead, it incorporates both visual information and language semantics for open-vocabulary classification and segmentation tasks.
Each initialized part query $\mathbf{p_i^0}$ is repeated $N_p$ times to accommodate multiple instances of the same part category within a single object (e.g., a bus having four wheels). 
Consequently, the part queries of the same part category and the same object are identical, e.g., $\mathbf{p^0}$ of bus1’s wheel1 and bus1’s wheel2.
On the other hand, the queries of the same category but different objects are different due to the distinct visual information from the objects, e.g., $\mathbf{p^0}$ of bus1’s wheel1 and bus2’s wheel1.
}
The initialized query is further refined by a MLLM to link the visual and text information between the object and its corresponding part, as detailed in the following subsection.

\subsection{MLLM-based Parsing}\label{subsec:MLLM_parsing}
\crv{
To parse the multi-granularity concepts embedded in $\mathbf{P^0} = \{\mathbf{p_i^0}\} $, we utilize PaliGemma 2~\cite{steiner2024paligemma2familyversatile}, a lightweight and state-of-the-art MLLM that takes the image $I$, the concatenated object-part queries $\mathbf{P^0}$, and prompt guidance as input and implements object-part parsing in our framework. 
From the prompt, the MLLM receives the object query $\mathbf{O}^L$ followed by part queries $\mathbf{P^0}$ and outputs refined part queries that integrate both object- and part-level information. This design enables the MLLM to leverage object-level context to infer part semantics, and also allows bidirectional information flow - from parts back to the object - during training (see Sec.~\ref{subsec:ablation} Object-Part Synergy).
}
Subsequently, the image tokens from SigLIP~\cite{steiner2024paligemma2familyversatile} and the prompt with queries $\mathbf{P^0}$ are provided to Gemma 2 model in a structured text prompt as follows: 
\vspace{-5pt}
\begin{minipage}{\textwidth}
\begin{lstlisting}[basicstyle=\scriptsize\itshape]
Please do object-part parsing on the image <img><img_tokens></img>. 

For each object, you will be given a list of object-part queries:
<obj_part>part_query1, part_query 2, ..., part_query n</obj_part>, 
please implement object-part parsing by refine the queries so that it can be used 
for later part category and mask prediction.

These are all the candidate object-part queries: 
    object 1 with parts <obj_part>part_query1, part_query 2, ..., 
        part_query n1</obj_part> ; 
    object 2 with parts <obj_part>part_query1, part_query 2, ..., 
        part_query n2</obj_part>; 
    ...
\end{lstlisting}
\end{minipage}

This stage processes object-part queries jointly and outputs part queries $\mathbf{P}$ integrated with visual information and object context. 
Note we utilize Gemma 2 as a feed-forward model, instead of utilizing auto-regressive generation to ensure a controlled output structure. 
$\mathbf{P}$ is obtained from the last hidden states of the corresponding input part queries. 
$\mathbf{P}$ will be used as input for a separate part decoder with same structure as the object decoder introduced in Sec.~\ref{subsec:obj_seg}.

\subsection{Implementation Details}
We employ a two-stage training strategy. 
In the first stage, we train the model with an object instance segmentation loss only:
\begin{equation}
    L^1 = \lambda_{\mathrm{cls}} \cdot L^\mathrm{obj}_\mathrm{cls} +  \lambda_\mathrm{bbox} \cdot L^\mathrm{obj} _\mathrm{bbox}  + \lambda_\mathrm{mask} \cdot L^\mathrm{obj}_\mathrm{mask}.
\end{equation}
$L^{\mathrm{obj}}_{\mathrm{cls}}$ is the focal loss~\cite{lin2017focalloss} on the prediction logits $\mathbf{S^O}$.
$L^{\mathrm{obj}}_{\mathrm{bbox}}$ is the L1 loss on predicted object bounding boxes $\mathbf{B^O}$.
$L^{\mathrm{obj}}_{\mathrm{mask}}$ is the combination of focal loss and dice loss~\cite{mille2016diceloss} on the predicted object masks $\mathbf{M^O}$.
In the second stage, joint object and part segmentation training is implemented with losses on both object and part predictions:
\begin{equation}
        L^2 = \lambda_{\mathrm{cls}} \cdot (L^{\mathrm{obj}}_{\mathrm{cls}} + L^{\mathrm{part}}_{\mathrm{cls}}) +  \lambda_{\mathrm{bbox}} \cdot ( L^{\mathrm{obj}}_{\mathrm{bbox}} + L^{\mathrm{part}}_{\mathrm{bbox}}) + \lambda_{\mathrm{mask}} \cdot (L^{\mathrm{obj}}_{\mathrm{mask}} + L^{\mathrm{part}}_{\mathrm{mask}}) .
\end{equation}
The loss functions on part segmentation are the same with the ones on object. The parameters of the Swin-L backbone and MaskDINO decoder are initialized with the pre-trained checkpoints from GLEE~\cite{wu2023GLEE}.
Following MaskDINO, the hyperparameters are set to $\lambda_{\mathrm{cls}} = 4, \lambda_{\mathrm{bbox}} = 2, \lambda_{\mathrm{mask}} = 5, L = 9$.
The number of repeated part queries $N_p = 3$.
The training is conducted on 4 x H200 GPUs with a batch size of 16.

\section{Experiments}

\begin{table}[tb]
  \centering
  \scriptsize
  \begin{tabular}{l@{\hskip 14pt}c@{\hskip 6pt}c@{\hskip 6pt}c @{\hskip 14pt}c@{\hskip 6pt}c@{\hskip 6pt}c @{\hskip 14pt}c@{\hskip 6pt}c@{\hskip 6pt}c @{\hskip 14pt}c@{\hskip 6pt}c@{\hskip 6pt}c }
  \toprule
  \multirow{2}{*}{Method} 
  & \multicolumn{3}{c}{\textbf{PPS-116}} 
  & \multicolumn{3}{c}{\phantom{obj}+\textit{INS}\phantom{obj}} 
  & \multicolumn{3}{c}{\phantom{obj}+\textit{INS}+\textit{PART}\phantom{obj}} 
  & \multicolumn{3}{c}{\color{gray}{PartImageNet}} \\
    
    & obj & part & AP 
    & obj & part & AP
    & obj & part & AP 
    & \color{gray}{obj} & \color{gray}{part} & \color{gray}{$AP$} \\
    
    \midrule
    VLPart 
    & -- & 4.5 & -- 
    & -- & -- & --
    & -- & -- & --
    & \color{gray}{--} & \color{gray}{29.7} & \color{gray}{--} \\
    
    PSALM$\dagger$ 
    & 31.6 & 8.27 & 13.4 
    & 48.0 & 10.7~({\color{cyan}{+2.4}}) & 18.9~({\color{green}{+5.5}}) 
    & 58.6 &  \underline{11.6}~({\color{cyan}{+3.3}}) & \underline{21.9}~({\color{green}{+8.5}}) 
    & \color{gray}{79.2} & \color{gray}{40.1} & \color{gray}{48.7} \\

    PartGLEE 
    & \underline{38.4} & \textbf{9.20} & \underline{15.6} 
    & \underline{58.7} & \underline{11.0}~(\color{cyan}{+1.8})  &  \underline{21.5}~({\color{green}{+5.9}}) 
    &  \underline{61.0} & 9.57~({\color{cyan}{+0.4}}) & 21.0~({\color{green}{+5.4}}) 
    & \color{gray}{\underline{81.4}} & \color{gray}{\underline{41.5}} & \color{gray}{\underline{50.4}} \\
    
    \OurMethod{}     
    & \textbf{44.5} & \underline{8.86} & \textbf{16.7} 
    & \textbf{60.5} & \textbf{11.4}~(\color{cyan}{+2.5}) & \textbf{22.3}~({\color{green}{{+5.6}}}) 
    & \textbf{62.8} & \textbf{16.4}~(\color{cyan}{+7.5}) & \textbf{26.7}~({\color{green}{+10.}}) 
    & \color{gray}{\textbf{83.9}} & \color{gray}{\textbf{49.2}} & \color{gray}{\textbf{56.9}} \\
    
    \bottomrule
  \end{tabular}
  \vspace{8pt}
  \caption{Cross-dataset experiment: \textbf{PascalPart-116}~(training) $\rightarrow$ PartImageNet~(evaluation) and in-domain experiment:  PartImageNet~(training) $\rightarrow$ PartImageNet~(evaluation). 
  We report object-level (obj), part-level (part), and overall ($AP$) mAP. 
  The best result is in \textbf{bold} and the second best one is in \underline{underline}.
  The notations "+\textit{INS}" and "+\textit{INS}+\textit{PART}" indicate additional training dataset for scalability. 
  \textcolor{green}{Green values} reflect relative $AP$ gains over the PPS-116 baseline; 
  \textcolor{cyan}{Cyan values} reflect relative mAP gains over the PPS-116 baseline. 
  \textcolor{gray}{Gray columns} shows in-domain performance.}
  \label{tab:cross_dataset_partimagenet}
\vspace{-10pt}
\end{table}

\begin{table}[tb]
  \centering
  \scriptsize
  \begin{tabular}{l@{\hskip 14pt}c@{\hskip 6pt}c@{\hskip 6pt}c @{\hskip 14pt}c@{\hskip 6pt}c@{\hskip 6pt}c @{\hskip 14pt}c@{\hskip 6pt}c@{\hskip 6pt}c @{\hskip 14pt}c@{\hskip 6pt}c@{\hskip 6pt}c }
    \toprule
    \multirow{2}{*}{Method}
    & \multicolumn{3}{c}{\textbf{PartImgNet}}
    & \multicolumn{3}{c}{\phantom{obj}+\textit{INS}\phantom{obj}}
    & \multicolumn{3}{c}{\phantom{obj}+\textit{INS}+\textit{PART}\phantom{obj}}
    & \multicolumn{3}{c}{\color{gray}{PPS-116}} \\
    
    & obj & part & AP
    & obj & part & AP
    & obj & part & AP
    & \color{gray}{obj} & \color{gray}{part} & \color{gray}{$AP$} \\
    
    \midrule
    PSALM
    & \underline{8.58} & 1.87 & 2.89
    & 20.0 & \underline{3.33}~(\textcolor{cyan}{+1.5}) & 5.87~(\textcolor{green}{+3.0})
    & 20.1 & \underline{3.58}~(\textcolor{cyan}{+1.7}) & 6.09~(\textcolor{green}{+3.2})
    & \color{gray}{48.7} &\color{gray}{13.2} & \color{gray}{18.6} \\
    
    PartGLEE
    & 8.00 & \underline{2.18} & \underline{3.06}
    & \underline{23.1} & 3.06~(\textcolor{cyan}{+0.9}) & \underline{6.17}~(\textcolor{green}{+3.1})
    & \underline{22.2} & 3.37~(\textcolor{cyan}{+1.2}) & \underline{6.33}~(\textcolor{green}{+3.3})
    & \underline{\color{gray}{53.2}} &  \underline{\color{gray}{14.5}} & \color{gray}{\underline{20.4}} \\
    
    \OurMethod{}
    & \textbf{9.57} & \textbf{2.20} & \textbf{3.32}
    & \textbf{23.3} & \textbf{3.64}~(\textcolor{cyan}{+1.4}) & \textbf{6.63}~(\textcolor{green}{+3.3})
    & \textbf{22.6} & \textbf{4.67}~(\textcolor{cyan}{+2.5}) & \textbf{7.39}~(\textcolor{green}{+4.1})
    & \textbf{\color{gray}{54.6}} & \textbf{\color{gray}{15.0}} & \textbf{\color{gray}{21.0}} \\
    
    \bottomrule
  \end{tabular}
  \vspace{8pt}
  \caption{Cross-dataset experiment: \textbf{PartImageNet}~(training) $\rightarrow$ PPS-116~(evaluation) and in-domain experiment: PPS-116~(training) $\rightarrow$ PPS-116~(evaluation).}
  \label{tab:cross_dataset_pps116}
  \vspace{-20pt}
\end{table}


\subsection{Cross-dataset Object-Part Instance Segmentation}\label{subsec:experiemnt_cross_dataset}
We conduct experiments to evaluate the cross-dataset generalization performance of \OurMethod{}, as well as baseline methods for the object-part instance segmentation task.  
\par\noindent\textbf{Experiment Setup.}
We follow the setup proposed in VLPart~\cite{sun2023vlpart} where each method is trained on one base dataset and evaluated on another unseen dataset, without finetuning. 
Two settings are implemented: Pascal-Part-116~\cite{wei2023ovpart} $\rightarrow$ PartImageNet~\cite{he2021partimagenet} and PartImageNet $\rightarrow$ Pascal-Part-116 (i.e., the model is  trained on Pascal-Part-116 and evaluated on PartImageNet, and vice versa).
We further evaluate the scalability of \OurMethod{} by integrating two additional sets of datasets into training, including object-level datasets \textit{INS}~(consisting of COCO~\cite{coco_common}, VisualGenome~\cite{krishna2016visualgenome} and LVIS~\cite{lvis}, with object annotations) and part-level datasets~(\textit{PART} consisting of ADE20K~\cite{zhou2017ade20k}, SA1B~\cite{sam_2023_ICCV} and PACO~\cite{ramanathan2023paco}, with object and part annotations).
Note the granularity of the part-level annotations across the datasets within \textit{PART} are different.  
The metric is $mAP_{mask}$ on the evaluation set of PartImageNet and Pascal-Part-116 dataset.

\par\noindent\textbf{Baseline methods.}
The existing methods for the OVPIS task include VLPart~\cite{sun2023vlpart} and PartGLEE~\cite{li2024partglee}. 
To extend the set of baselines for comparison, we further adapt PSALM~\cite{zhang2025psalm}, a state-of-the-art LLM-based 2D object-level segmentation method by extending the LLM mask tokens with learnable part queries for object-part parsing and part segmentation.
The adapted PSALM is denoted as PSALM$\dagger$.
\par\noindent\textbf{Cross-dataset and in-domain evaluation on PartImageNet}.
As shown in Tab.~\ref{tab:cross_dataset_partimagenet}, \OurMethod{} achieves the best performance of object-part instance segmentation in both cross-dataset and in-domain settings (i.e.,trained with Pascal-Part-116 and evaluated on PartImageNet). \OurMethod{} surpasses PartGLEE by $1.1\%$ and PSALM$\dagger$ by $3.3\%$ in mAP on object-part instance segmentation. 
Our experiments further show that \OurMethod{} has better scalability with additional training datasets containing part-level annotations.
Trained on Pascal-Part-116+\textit{INS}, all methods achieve similar performance gains in both part-level mAP and overall AP.
However, when the training set is extended with additional part-level datasets~(Pascal-Part-116 + \textit{INS} + \textit{PART}) our approach achieves a significant performance boost in both part-level mAP~($\color{cyan}{+7.5}$) and overall AP~($\color{green}{+10.0}$). In contrast, the performance gain of PartGLEE in part-level segmentation drops~($\color{green}{+5.9} \rightarrow \color{green}{+5.4}$) compared to the Pascal-Part-116+\textit{INS} setting, mainly due to lacking the object-part hierarchy context during part parsing phase, as illustrated in Sec.~\ref{subsec:part_query}.
\par\noindent\textbf{Cross-dataset and in-domain evaluation on Pascal-Part-116}. 
As shown in Tab.~\ref{tab:cross_dataset_pps116}, training the model on PartImageNet and implementing evaluation on Pascal-Part-116 is more challenging than the previous condition for all the evaluated methods. Indeed, the latter dateset contains multiple novel object categories and finer-granularity parts than the former. 
\OurMethod{} achieves the best performance in both cross-dataset and in-domain object-part segmentation on Pascal-Part-116.
The experiments also shows the advantage of \OurMethod{} in scalability, especially in the setting with PartImageNet+ \textit{INS} + \textit{PART} as training dataset~({\color{green}{+4.1}} over {\color{green}{+3.2}} and {\color{green}{+3.3}})
 when trained only on PartImageNet dataset. Here, \OurMethod{} performs better than both baselines PartGLEE and PSALM$\dagger$. 
\par
\begin{figure}
\begin{center}
\includegraphics[width=0.95\linewidth]{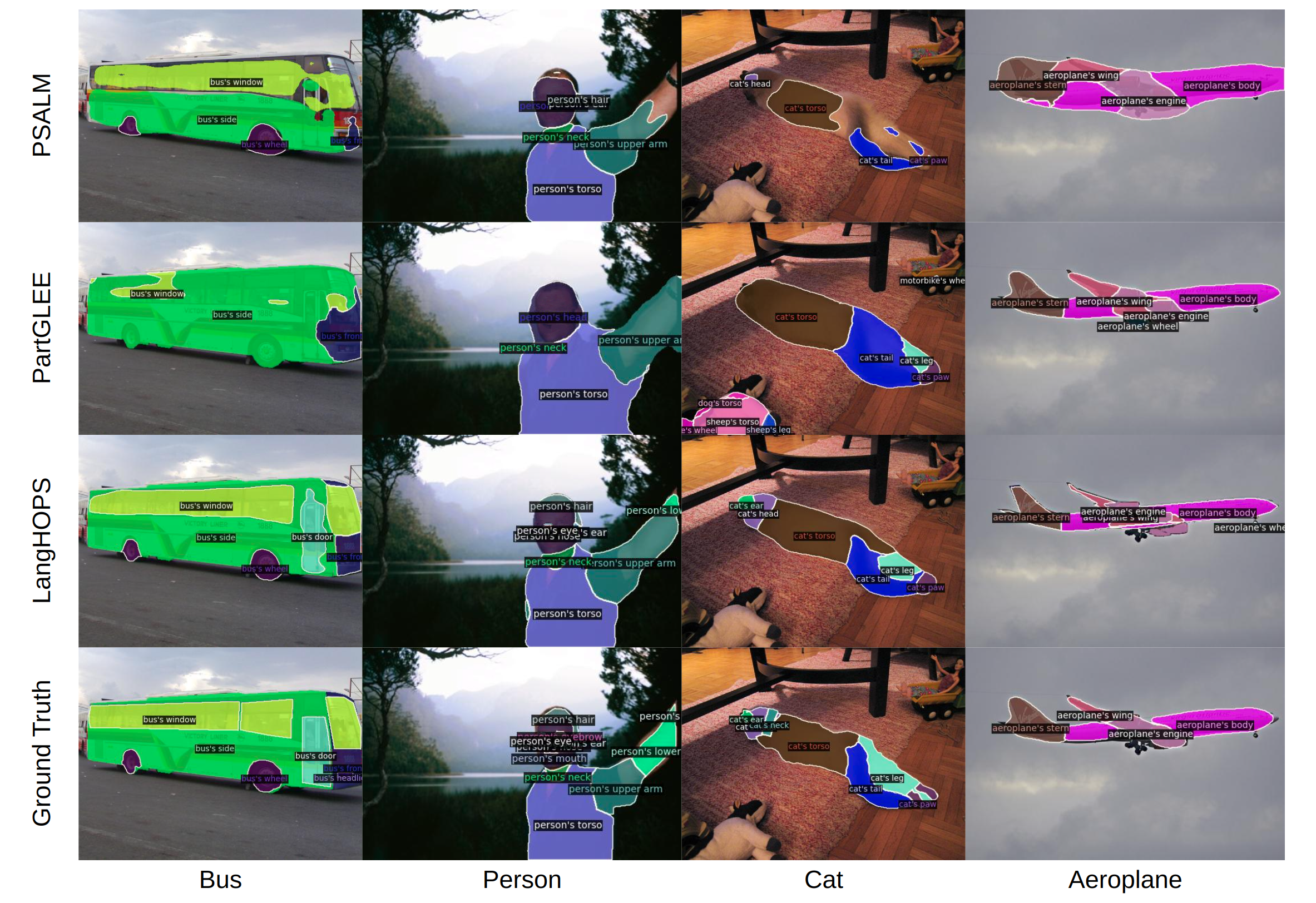}
      \vspace{-5pt}
    \caption{Qualitative results of part-level segmentation if \OurMethod{} and baselines.} \label{fig:qualitative}
\end{center}
\vspace{-14pt}
\end{figure}

\noindent\textbf{Qualitative Results} are shown in Fig.~\ref{fig:qualitative} in the setting of PartImageNet + \textit{INS} + \textit{PART} $\rightarrow$ Pascal-Part-116. 
As the figure shows, \OurMethod{} achieves more accurate part segmentation than other baselines. 
Importantly, segmentation results on "person" and cat demonstrates \OurMethod{}'s superior generalization performance to finer part granularity in the cross-dataset condition.

\subsection{Zero-shot Part Segmentation}
We further carry out experiments on the OV-Part benchmark~\cite{wei2023ovpart} and PartImageNet dataset~\cite{he2021partimagenet} for the zero-shot segmentation. 
One must note that this benchmark is evaluating open-vocabulary semantic segmentation of parts, which is not the core application of \OurMethod{}. The metric used is the harmonic mean of intersection-over-union (hIoU), for both seen and unseen categories~\cite{xian2019semantic_project_zero_label}:
\begin{equation}
    hIOU = \frac{2 \cdot mIoU_{\mathrm{seen}}\cdot mIoU_{\mathrm{unseen}}}{mIoU_{\mathrm{seen}} +  mIoU_{\mathrm{unseen}}}.
\end{equation}
As shown in Tab.~\ref{tab:zero_shot_ov_parts}, \OurMethod{} achieves the best performance on Pascal-Part-116 and PartImageNet datasets, and reaches second-best performance on ADE20K-234 dataset, achieving competitive performance with PartCATSeg~\cite{partcatseg}.
\OurMethod{} obtains the highest $mIoU_{\mathrm{seen}}$ on all three datasets, demonstrating the superior generalization ability to unseen object and part categories. 
Noticeably, our method is designed for open-vocabulary object-part instance segmentation while most others, including PartCATSeg~\cite{partcatseg}, are designed specifically for the OV-Part benchmark~(semantic part segmentation). 
\crv{
Directly evaluating \OurMethod{}{} in the semantic segmentation still leads to superior performance (hIOU) in PPS-116 and PartImageNet datasets, showing its great potential.}
\begin{table}[tb]
  \centering
  \scriptsize
  \begin{tabular}{l ccc ccc ccc }
  \toprule
     \multirow{2}{*}{Method} & \multicolumn{3}{c}{PPS-116~\cite{wei2023ovpart}} & \multicolumn{3}{c}{PartImageNet~\cite{he2021partimagenet}} & \multicolumn{3}{c}{ADE20K~\cite{wei2023ovpart}}  \\
     & seen & unseen & ${hIoU}$ & seen & unseen & ${hIoU}$ & seen & unseen & ${hIoU}$ \\
     \midrule
    VLPart~\cite{sun2023vlpart} & 42.6 & 18.7 & 26.0 & * & * & * & * & * & * \\
    ZSSeg+~\cite{xu2023zsseg} & 54.4 & 19.0 & 28.2 & * & * & * & 43.2 & 27.8 & 33.9 \\
    CLIPSeg~\cite{Luddecke2024clipseg, wei2023ovpart} & 48.9 & 27.5 & 35.2 & 53.9 & 37.2 & 44.0 & 38.2 & 30.9 & 34.2 \\
    CAT-Seg~\cite{cho2024catseg, wei2023ovpart} & 43.8 & 27.7 & 33.9 & 47.3 & 35.1 & 40.3 & 33.8 & 25.9 & 29.3 \\
    PartCLIPSeg~\cite{choi2024partclipseg} & 50.0 & 31.7 & 38.8 & 56.3 &  51.7 & 53.9 & 38.4 & 38.8 & 38.6 \\
    PartGLEE~\cite{li2024partglee} & 57.4 & 27.4 & 37.1 & * & * & * &  \underline{51.3} & 35.3 & 41.8 \\
    PartCATSeg~\cite{partcatseg} & \underline{57.5} & \underline{44.9} &  \underline{50.4} & \textbf{73.8} & \underline{71.5} & \underline{72.7} &  \textbf{53.1} & \underline{47.2} & \textbf{50.0} \\
    \OurMethod{} & \textbf{59.2} & \textbf{46.5} & \textbf{52.1} & \underline{71.9} & \textbf{73.7} & \textbf{72.8} & 49.3 & \textbf{49.7} & \underline{49.5} \\ 
    \bottomrule
  \end{tabular}
      \vspace{5pt}
    \caption{h-IoU. Zero-shot evaluation on PPS-116, PartImageNet and ADE20K. } \label{tab:zero_shot_ov_parts}
    \vspace{-10pt}
\end{table}

\subsection{Ablation Study}\label{subsec:ablation}
\begin{table}[ht]
\centering
\scriptsize
\renewcommand{\arraystretch}{1.2}
\begin{tabularx}{\textwidth}{c YY YY YY}
\toprule
Training setup & \multicolumn{2}{c}{Obj Seg} & \multicolumn{2}{c}{Detached Obj-Part Seg} & \multicolumn{2}{c}{Obj-Part Seg} \\
Eval Dataset & Obj & Part & Obj & Part & Obj & Part \\
\midrule
PPS-116 & 25.8 & 0.00 & 25.2 & 9.66 & 26.2 & 10.3 \\
PartImageNet & 67.9 & 2.08 & 62.9 & 13.2 & 68.3 & 14.9 \\
\bottomrule
\end{tabularx}
\vspace{8pt}
\caption{mAP of object and part instance segmentation. Ablations on Object-Part Synergy.}
\label{tab:object_part_synergy}
\vspace{-15pt}
\end{table}

\begin{wraptable}{r}{0.3\textwidth} 
\vspace{-20pt}
  \centering
  \small
  \scriptsize
  \begin{tabular}{l c c}
    \toprule
    Setting & Detached & Obj-Part Seg \\
    \midrule
    obj  & 0.76 & 0.82 \\
    part  & 0.58 & 0.67 \\
    \bottomrule
  \end{tabular}
  \caption{Attention score.}
  \label{tab:attention_score}
  \vspace{-15pt}
\end{wraptable}
Ablation studies further demonstrate the effectiveness of \OurMethod{}. 
\par\noindent\textbf{Object-Part Synergy}.
To showcase the object-part synergy enabled by \OurMethod{} (i.e., a performance improvement from joint training of object and part instance segmentation) we reported the our performances in following training setups:
a) "Obj Seg": \OurMethod{} is trained only with the loss of object instance segmentation;  
b) "Detached Obj-Part Seg": \OurMethod{} is trained using losses of both object and part instance segmentation. However, the gradient flow coming from "MLLM-based parsing" module is interrupted, meaning that the gradients of $\mathbf{P}$ from the part segmentation loss will not directly propagate to object queries $\mathbf{O}^L$. One can note that object and part segmentation will still affect each other indirectly since both tasks use the same dense image features $\mathbf{X}$.
c) "Obj-Part Seg": This setup allows a joint training of object and part instance segmentation without gradient flow cut.
As show in Tab.~\ref{tab:object_part_synergy}, in "Obj Seg" setting, the mAP of part segmentation performance is near $0$, as the loss of part segmentation is not used. 
Compared to "Obj Seg", the performance of object segmentation of "Detached Obj-Part Seg" drops $0.6\%$ on PascalPart116 dataset and more significantly on PartImageNet dataset by $5.0\%$, due to the absence of the gradient flow by the MLLM-based parsing.
In contrast, \OurMethod{} shows improved object segmentation performance in "Obj-Part Seg" than "Obj Seg", and gains significant boost in both object~(by $\textbf{5.4\%}$) and part segmentation~(by $\textbf{1.7\%}$) over "Detached Obj-Part Seg". This demonstrates that the proposed MLLM-based object-part parsing enables beneficial synergy effect in both cross-dataset and in-dataset conditions.
\crv{
We further investigate the object-part synergy mechanism by reporting the average attention score. 
The average attention score is calculated by summing attention scores of true positive predictions inside the ground truth masks $M$, divided by the area of the masks. 
The attention is the normalized cos similarity between object queries and the dense features of the final layer of the object/part decoder. 
\begin{equation}
    {S}_{a} = \sum_{u\in M} \frac{1 + cos(f_{u}, p_M)}{2\cdot |M|},
\end{equation}
where $u$ is the pixel within the ground truth mask $M$, $(f_{u}$ is the mapped feature for segmentation and $p_M$ is the refined object/part query of the predicted instance matched to the ground truth instance.
The score shows the amount of attention correctly assigned by the model to the ground truth area, and is in the range of [0, 1].
In the setting of PPS116+INS+PART -> PartImageNet, as shown in Tab.~\ref{tab:attention_score}, compared to the "detached object-part seg.", the synergized object-part segmentation leads to higher attention scores for both object and part segmentation, proving strong evidence of the synergy between both segmentation tasks.}

\par\noindent\textbf{Effect of MLLM-based Parsing}.
We implement an ablation study to demonstrate the effectiveness of the MLLM-based Object-Part Parsing module by replacing it with a Q-Former. The Q-Former takes the object queries $\mathbf{O}$ as key and value, and hierarchical part queries $\mathbf{P^0}$ as query. In the end, the Q-former-based module outputs part queries $\mathbf{P^{Q}}$ for part segmentation purposes.
As shown in Tab.~\ref{tab:ablation_modules}, the ablated version, denoted as "w/o MLLM" shows inferior performances with both PartImageNet and PPS116 datasets, demonstrating the effectiveness of the MLLM module in object-part parsing. 

\paragraph{Ablation on two-stage.} 
We also provide an ablation study on the training strategy of the model.
Two-stage refers to firstly training the model on object segmentation and secondly training it on object-part segmentation.
One-stage means we directly train the model on object-part segmentation from scratch.
Tab.~\ref{tab:ablation_training} shows the model trained with the two-stage strategy achieves better cross-dataset performance, though its in-domain performance is inferior compared to one-stage.
\begin{table}[tb]
  \centering
  \small
  \begin{tabular}{l@{\hskip 14pt}c@{\hskip 6pt}c@{\hskip 6pt}c @{\hskip 14pt}c@{\hskip 6pt}c@{\hskip 6pt}c @{\hskip 14pt}c@{\hskip 6pt}c@{\hskip 6pt}c @{\hskip 14pt}c@{\hskip 6pt}c@{\hskip 6pt}c }
  \toprule
  \multirow{2}{*}{Method} 
  & \multicolumn{3}{c}{\textbf{PPS-116}} 
  & \multicolumn{3}{c}{\phantom{obj}+\textit{INS}\phantom{obj}} 
  & \multicolumn{3}{c}{\phantom{obj}+\textit{INS}+\textit{PART}\phantom{obj}} 
  & \multicolumn{3}{c}{\color{gray}{PartImageNet}} \\
    
    & obj & part & AP 
    & obj & part & AP
    & obj & part & AP 
    & \color{gray}{obj} & \color{gray}{part} & \color{gray}{$AP$} \\
    \midrule
    One-Stage         
    & {40.6} & {8.50} & {15.6} 
    & {57.8} & {10.6} & {21.1} 
    & {60.2} & {15.5} & {25.4} 
    & \color{gray}{{84.6}} & \color{gray}{{51.2}} & \color{gray}{{58.6}} \\
    
    Two-Stage      
    & {44.5} & {8.86} & {16.7} 
    & {60.5} & {11.4} & {22.3} 
    & {62.8} & {16.4} & {26.7} 
    & \color{gray}{{83.9}} & \color{gray}{{49.2}} & \color{gray}{{56.9}} \\
    
    \bottomrule
  \end{tabular}
  \vspace{8pt}
  \caption{Ablations on training strategy in the cross-dataset setting of \textbf{PascalPart-116}~(training) $\rightarrow$ PartImageNet~(evaluation).}
  \label{tab:ablation_training}
\vspace{-20pt}
\end{table}

\begin{wraptable}{r}{0.4\textwidth} 
  \centering
  \small
  \scriptsize
  \begin{tabular}{l c c}
    \toprule
    Module & PartImageNet & PPS116 \\
    \midrule
    w/o MLLM & 23.2 & 18.4 \\
    w/o hierarchy & 22.5 & 19.1 \\
    \OurMethod{} & 26.7 & 19.8 \\
    \bottomrule
  \end{tabular}
  \caption{mAP on object-part instance segmentation in the cross-dataset setting. Ablations on architecture design.}
  \label{tab:ablation_modules}
  \vspace{-10pt}
\end{wraptable}

\par\noindent\textbf{Effect of Language-grounded Hierarchies}. 
To investigate the effectiveness of the language-space-aligned object-part hierarchies, we conduct an ablation study by replacing the representation proposed in Sec.~\ref{subsec:MLLM_parsing} with $N$ learnable queries, denoted as "w/o hierarchy" in Tab.~\ref{tab:ablation_modules}.
Specifically, we initialize $N$ learnable queries and concatenate them to each object query $\mathbf{o}^L$ from $\mathbf{O}^L$ to form the initial part queries $\mathbf{P^N}$.
Subsequently, the MLLM uses the $\mathbf{O}^L$, $\mathbf{P^N}$ and as input, and outputs parsed part queries that are forwarded to the part decoder for final part segmentation. 
As shown in Tab.~\ref{tab:ablation_modules}, by leveraging hierarchies between object and parts, and formulating part queries within language space, \OurMethod{} achieves better performance than the ablated version with learnable initial queries in both datasets.



\subsection{Limitation and Future Work}
As shown in the supplementary material (Section~\ref{subsec:suppl_computation_cost}), the computational cost of \OurMethod{} is nontrivial compared to the baselines, primarily due to the integration of the MLLM for object–part parsing. 
Improving efficiency is essential for deploying \OurMethod{} in real-time or on-board computer vision and robotics applications. 
In addition, the training datasets~\cite{chen2014pascalpart, he2021partimagenet, coco_common, krishna2016visualgenome, lvis} used in this work mainly contain common object and part categories, which may not fully cover all potential application scenarios. 
Therefore, additional datasets with task-specific annotations may still be required for fine-tuning in specialized cases (e.g., interactable articulated objects for robotic manipulation), even though \OurMethod{} demonstrates strong generalization capabilities compared to existing baselines.
Additionally, as 2D-to-3D lifting~\cite{langsplat, panoptic_lift, miao2024volumetric, zhang2024styletransfer} is increasingly popular, leveraging \OurMethod{}{} for 3D computer vision tasks~\cite{miao2024scenegraphloc, halacheva2025articulate3d, zhang2025inverse_render, zhang2024glorieslam, Sandstrom_2025_CVPR} is also a promising future direction.

\section{Conclusion}
We propose a new method \OurMethod{} that performs Open-vocabulary Part Instance Segmentation through hierarchical modeling in language space. Using language-grounded hierarchies improves both the context awareness and the accuracy of object-part parsing. In experiments, we show that \OurMethod{} performs notably better than existing state-of-the-art methods across multiple benchmark settings. Notably, our method achieves significant improvements in in-domain and cross-dataset object-part instance segmentation, where we outperform existing state-of-the-art approaches by $\textbf{5.5\%}$ AP. \OurMethod{} further achieves the best $mIOU$ on unseen object-parts in OVPIS tasks, on all PartImageNet, PascalPart-116 and ADE20K datasets, consequently demonstrating strong generalization ability in unseen object and part categories. 
In conclusion, \OurMethod{} establishes a novel foundation for Open-vocabulary Part Segmentation and highlights the potential of MLLM-based methods for fine-grained visual understanding, with the aim of encouraging further research into scalable language-driven approaches for structured scene parsing.

\section{Acknowledgement}
This research was funded by Toyota Motor Europe and the Ministry of Education and Science of Bulgaria (support for INSAIT, part of the Bulgarian National Roadmap for Research Infrastructure).




{
    \small
    \bibliographystyle{ieeenat_fullname}
    \bibliography{main}
}


\appendix

\clearpage
\section{Technical Appendices and Supplementary Material}
This section provides additional visualization and ablation studies.

\subsection{Visualization}
\begin{figure}[ht]
  \centering
  \includegraphics[width=0.99\linewidth]{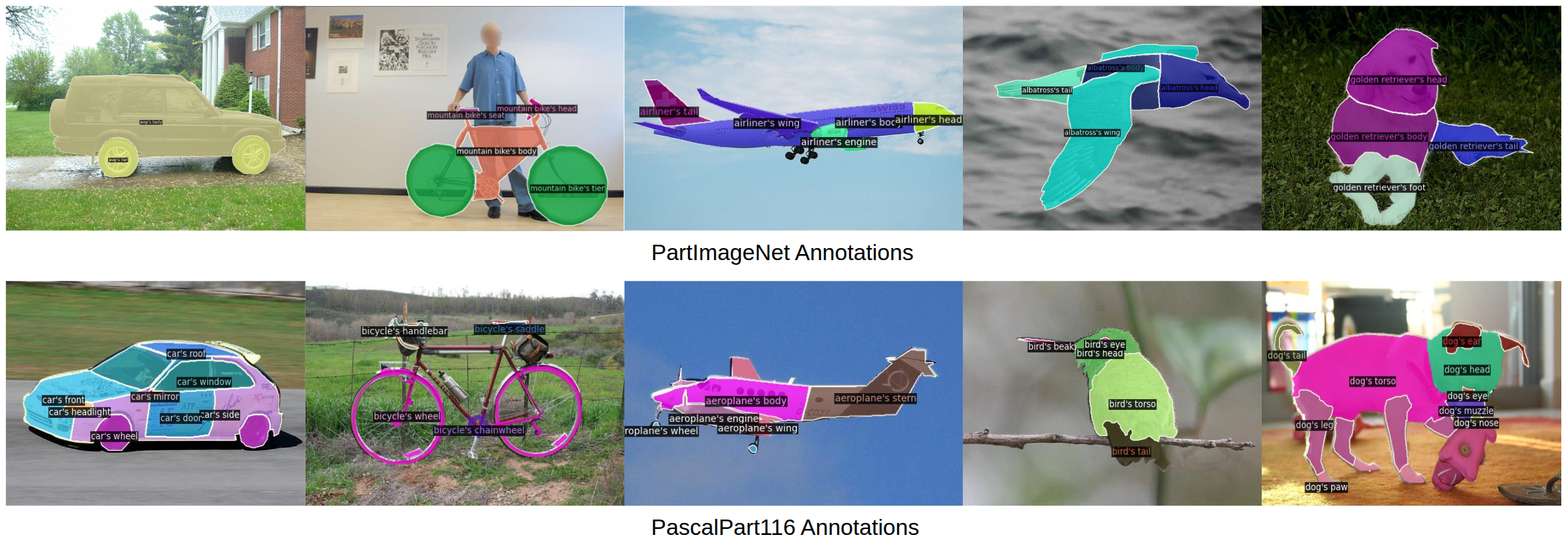} 
  \caption{
  Visualization on Annotations of PartImageNet and PascalPart116 datasets. 
  }
  \label{fig:different_granularity}
\end{figure}
\begin{figure}[ht]
  \centering
  \includegraphics[width=0.99\linewidth]{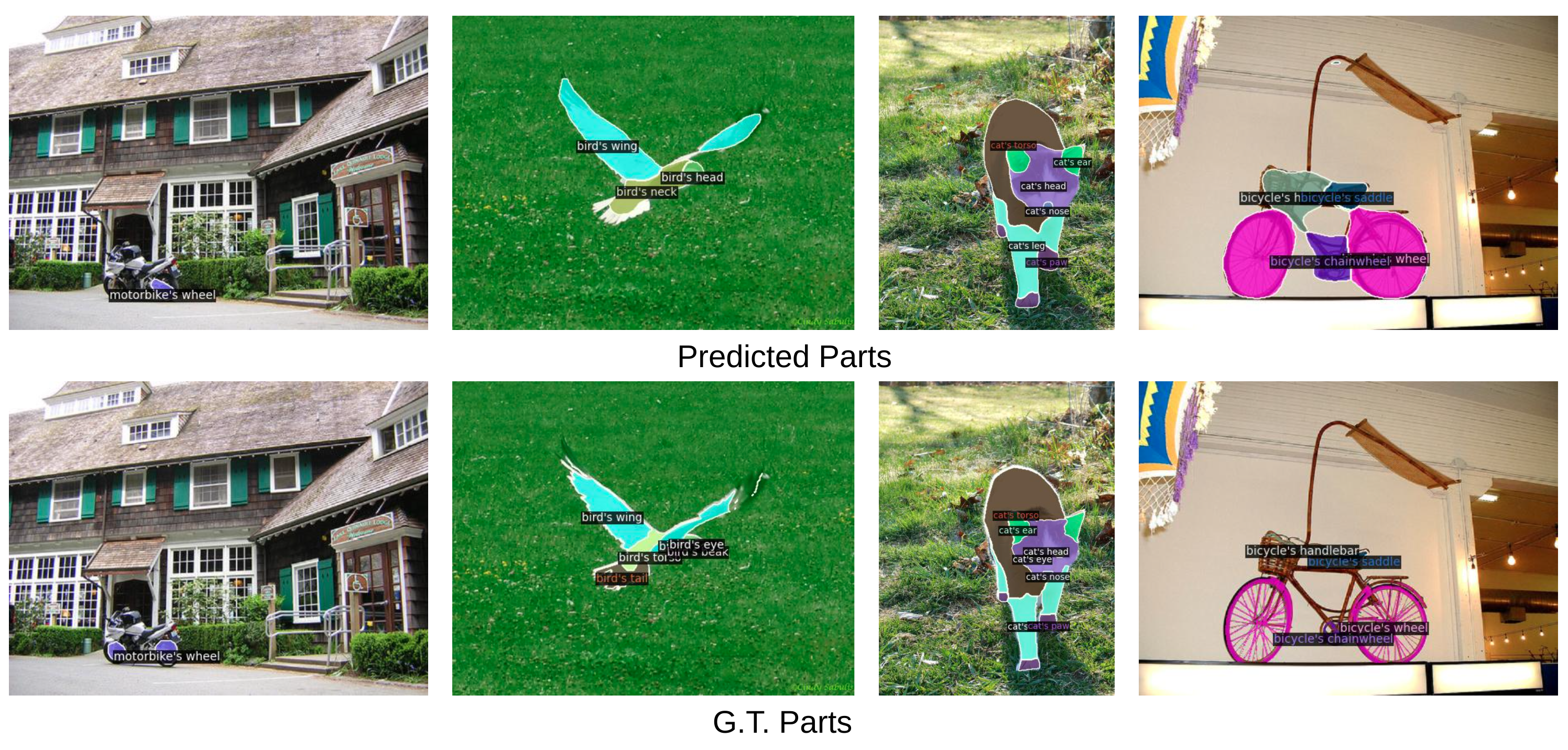} 
  \caption{
  Failure cases of \OurMethod{} in the cross-dataset setting of \textbf{PartImageNet}+\textit{INS}+\textit{PART}~(training) $\rightarrow$ ~PPS-116(evaluation). 
  }
  \label{fig:failure_cases}
\end{figure}

\paragraph{Granularity Difference Across Dataset.}In Fig.~\ref{fig:different_granularity}, we provide additional visualizations of the annotations of PartImageNet and PascalPart116 datasets. 
The figure shows that the two datasets provide annotations of parts in different granularity. 
Generally, PascalPart116 has finer part definition and thus it is more challenging to implement part segmentation on PascalPart116 than on PartImageNet, which explains that in both cross-dataset and in-domain settings, \OurMethod{} and baselines achieve less $mAP$ on PascalPart116 than on PartImageNet.
%
\paragraph{Failure Cases.} We further provide failure cases of \OurMethod{} in the cross-dataset setting. 
As shown in Fig.~\ref{fig:failure_cases}, \OurMethod{} can fail in several cases: 
\begin{itemize}
    \item when the object is distant to the camera and has small area in the image, \OurMethod{} may not be able to detect all the parts~(one motorbike's wheel missing);
    \item in the cross-dataset setting, \OurMethod{} have difficulties in generalizing to some novel parts which it has not see during training~(bird's eye, cat's eye). As shown in Fig.~\ref{fig:different_granularity}, the training dataset~(PartImagenet) only contain annotations of animal's head and no annotation of eyes.
    \item when the training and evaluation dataset have different annotation styles, the trained model tends to predict the part segmentation in the style of training dataset~(bicycle's wheel, all the pixels within the wheel circle).
\end{itemize} 

\subsection{Robustness Analysis}
\paragraph{Statistical Robustness of Evaluation.} 
\crv{
We conduct repetitive experiments the same in Sec.~\ref{subsec:experiemnt_cross_dataset} with 3 different random seeds.
The average and standard deviation are calculated and reported in Tab.~\ref{tab:experiment_std_cross_dataset}.
The table shows the statistical stability of the cross-dataset evaluation and verifies the superiority of the proposed \OurMethod{}{} over the baseline. }
\begin{table}[ht]
\centering
\scriptsize
\begin{subtable}{\linewidth}
\centering
  \begin{tabular}{l@{\hskip 14pt}c@{\hskip 6pt}c@{\hskip 6pt}c @{\hskip 14pt}c@{\hskip 6pt}c@{\hskip 6pt}c @{\hskip 14pt}c@{\hskip 6pt}c@{\hskip 6pt}c @{\hskip 14pt}c@{\hskip 6pt}c@{\hskip 6pt}c }
  \toprule
  \multirow{2}{*}{Method} 
  & \multicolumn{3}{c}{\textbf{PPS-116}} 
  & \multicolumn{3}{c}{\phantom{obj}+\textit{INS}\phantom{obj}} 
  & \multicolumn{3}{c}{\phantom{obj}+\textit{INS}+\textit{PART}\phantom{obj}} \\
    
    & obj & part & AP 
    & obj & part & AP
    & obj & part & AP \\
    
    \midrule
    \textbf{PartGLEE} 
    & 38.4±0.5 & 8.61±0.46 & 15.2±0.5 
    & 57.6±1.8 & 11.3±0.2 & 21.5±0.6
    & 60.1±0.9 & 10.5±0.7 & 21.5±0.7 \\
    \textbf{LangHOPS}
    & 48.7±3.3 & 8.89±0.24 & 17.7±0.9
    & 60.9±0.7 & 12.1±1.0 & 22.9±1.0
    & 63.7±1.4 & 16.6±0.3 & 27.0±0.5 \\
    \bottomrule
  \end{tabular}\caption{PPS-116 $\rightarrow$ PartImageNet}
\end{subtable}
\begin{subtable}{\linewidth}
\centering
  \begin{tabular}{l@{\hskip 14pt}c@{\hskip 6pt}c@{\hskip 6pt}c @{\hskip 14pt}c@{\hskip 6pt}c@{\hskip 6pt}c @{\hskip 14pt}c@{\hskip 6pt}c@{\hskip 6pt}c @{\hskip 14pt}c@{\hskip 6pt}c@{\hskip 6pt}c }
\toprule
  \multirow{2}{*}{Method} 
  & \multicolumn{3}{c}{\textbf{PPS-116}} 
  & \multicolumn{3}{c}{\phantom{obj}+\textit{INS}\phantom{obj}} 
  & \multicolumn{3}{c}{\phantom{obj}+\textit{INS}+\textit{PART}\phantom{obj}} \\
    & obj & part & AP 
    & obj & part & AP
    & obj & part & AP \\
\midrule
\textbf{PartGLEE} 

    & 8.53±0.52 & 2.05±0.09 & 3.04±0.16
    & 23.3±0.1 & 3.10±0.16 & 6.17±0.16
    & 22.5±0.4 & 3.60±0.24 & 6.46±0.26\\
\textbf{LangHOPS}
    & 11.0±1.0 & 2.17±0.03 & 3.50±0.18	
    & 22.9±0.7 & 3.68±0.11 & 6.59±0.19
    & 23.2±0.5 & 4.51±0.25 & 7.34±0.28 \\
\bottomrule
\end{tabular}\caption{PartImageNet $\rightarrow$ PPS-116}
\end{subtable}
\caption{Evaluation of PartGLEE and LangHOPS with mean ± standard deviation over 3 runs with different random seeds in the cross-dataset settings .}\label{tab:experiment_std_cross_dataset}
\end{table}

\paragraph{Robustness to Prompt Formulation.}
\crv{
We conducted two ablation studies on the ordering and wording of the structured input prompts to assess the robustness of our method to prompt formulation.
\textbf{(a) robustness to prompt ordering}: We randomly shuffled (i) the order of object queries, and (ii) the order of part queries within each object, multiple times during inference. For instance, object 3 may appear before object 1, or part queries within an object may be permuted (e.g., "part 9, part 4, part 6"). 
As shown in Tab.~\ref{tab:ablation_query_ordering}, our method remains highly stable across these permutations, with minimal performance degradation, demonstrating robustness to input ordering.
\textbf{(b) robustness to wording}: We further test the model's robustness to unseen part names by replacing the subset (from $0$ to $100\%$) of the original part category names with GPT-4o-generated synonyms (e.g., "foot" → "leg"). As shown in Tab.~\ref{tab:ablation_synonym}, LangHOPS significantly outperforms PartGLEE under increasing synonym replacement ratios, indicating strong generalization to semantically similar but unseen phrasing. Note that synonym substitutions may introduce granularity mismatches with the dataset’s ground-truth annotations (e.g., "leg" may exclude "paw" in the ground truth for “foot”), which partially explains the observed performance drop.
}

\begin{table}[t]
\centering
\scriptsize
\setlength{\tabcolsep}{4pt}
\renewcommand{\arraystretch}{1.2}
\begin{tabular}{lcccccccccccc}
\toprule
\textbf{Method} & \multicolumn{3}{c}{\textbf{PPS-116}} & \multicolumn{3}{c}{\textbf{+INS}} & \multicolumn{3}{c}{\textbf{+INS+PART}} \\
\cmidrule(lr){2-4}\cmidrule(lr){5-7}\cmidrule(lr){8-10}\cmidrule(lr){11-13}
 & obj & part & AP & obj & part & AP & obj & part & AP \\
\midrule
Shuffling of Object & 47.8$\pm$2.7 & 8.57$\pm$0.36 & 17.1$\pm$0.9 & 61.1$\pm$0.8 & 11.7$\pm$0.8 & 22.6$\pm$0.8 & 65.1$\pm$0.9 & 15.8$\pm$0.3 & 26.7$\pm$0.4 \\
Shuffling of Part & 46.9$\pm$2.4 & 9.08$\pm$0.33 & 17.5$\pm$0.8 & 58.8$\pm$1.1 & 13.6$\pm$0.9 & 23.6$\pm$0.9 & 64.2$\pm$1.6 & 16.9$\pm$0.3 & 27.4$\pm$0.6 \\
No shuffling & 48.7$\pm$3.3 & 8.89$\pm$0.24 & 17.7$\pm$0.9 & 60.9$\pm$0.7 & 12.1$\pm$1.0 & 22.9$\pm$0.97 & 63.7$\pm$1.4 & 16.6$\pm$0.3 & 27.0$\pm$0.5 \\
\bottomrule
\end{tabular}
\vspace{5pt}
\caption{Ablations on the ordering of the object and part queries – PPS116$\rightarrow$PartImageNet.}
\label{tab:ablation_query_ordering}
\end{table}

\begin{table}[tb]
  \centering
  \scriptsize
  \begin{tabular}{l@{\hskip 10pt}c@{\hskip 4pt}c@{\hskip 10pt}c@{\hskip 4pt}c@{\hskip 10pt}c@{\hskip 4pt}c@{\hskip 10pt}c@{\hskip 4pt}c@{\hskip 10pt}c@{\hskip 4pt}c@{\hskip 10pt}c@{\hskip 4pt}c}
  \toprule
  Method 
  & \multicolumn{2}{c}{$0\%$} 
  & \multicolumn{2}{c}{$25\%$} 
  & \multicolumn{2}{c}{$50\%$} 
  & \multicolumn{2}{c}{$75\%$} 
  & \multicolumn{2}{c}{$100\%$} \\
  
   & part & AP 
   & part & AP
   & part & AP 
   & part & AP
   & part & AP \\
  \midrule
  PartGLEE 
  & 11.2 & 21.8 
  & 9.3 & 20.3 
  & 8.6 & 19.7 
  & 6.6 & 18.2 
  & 5.1 & 17.0 \\
  
  LangHOPS
  & 17.0 & 27.1 
  & 16.2 & 26.5 
  & 16.5 & 26.7 
  & 14.6 & 25.3 
  & 12.7 & 23.8 \\
  \bottomrule
  \end{tabular}
  \vspace{5pt}
  \caption{Ablation on the robustness to input part category names. 
  PPS116 + INS + PART $\rightarrow$ PartImageNet. 
  Different percentages of part category names replaced with GPT-4o generated synonyms.}
  \label{tab:ablation_synonym}
\end{table}

\paragraph{Robustness to Noisy Hierarchy.}
\crv{
We test on the common OVS setting using clean object-part hierarchies, but believe in the value of closing the gap towards a noisy real-world deployment. 
To evaluate the robustness of LangHOPS to noisy or automatically mined hierarchies, we replace a portion of the clean object-part taxonomy with GPT-4o-generated object-part hierarchies. 
These auto-mined hierarchies are constructed solely from the object category names and may introduce ambiguity, inconsistency, or irrelevant parts. 
In Tab.~\ref{tab:abation_noisy_hierarchy}, we report performance under the varying noise hierarchies as the input prompt while the remaining the clean dataset annotations for evaluation. 
We observe that:
\OurMethod{}{} consistently outperforms PartGLEE across all noise levels;
LangHOPS degrades more gracefully as noise increases, maintaining reasonable AP even when the hierarchies are noisy;
The performance gap widens especially at high noise levels, demonstrating LangHOPS's stronger resilience to imperfect or automatically mined hierarchies.
Please note that the auto-generated hierarchies are often inconsistent with the ground truth annotations in the dataset, leading to lower evaluation metrics. 
Overall, developing evaluation protocols for adaptive, task-specific hierarchies remains an open problem and a promising direction for future benchmark design.}
\begin{table}[t]
\centering
\scriptsize
\setlength{\tabcolsep}{5pt}
\renewcommand{\arraystretch}{1.1}
\begin{tabular}{lcccccccccc}
\toprule
\textbf{Method} & \textbf{0\%} & - & \textbf{25\%} & - & \textbf{50\%} & - & \textbf{75\%} & - & \textbf{100\%} & - \\
 & part & AP & part & AP & part & AP & part & AP & part & AP \\
\midrule
PartGLEE  & 11.2 & 21.8 & 10.3 & 21.1 & 9.8 & 20.7 & 8.2 & 19.4 & 3.6 & 15.8 \\
LangHOPS  & 17.0 & 27.1 & 13.1 & 24.1 & 12.4 & 23.5 & 8.8 & 20.7 & 6.7 & 19.1 \\
\bottomrule
\end{tabular}
\vspace{5pt}
\caption{\textbf{Ablations on the noisy hierarchy construction.}
Different percentages of obj–part hierarchies from the dataset are replaced with GPT-4o generated ones.}
\label{tab:abation_noisy_hierarchy}
\end{table}

\subsection{Additional Ablation study}
\paragraph{Ablation on $N_p$.}
We further provide ablation study on the number of repeated part queries for each object $N_p$ in the cross-dataset setting of \textbf{PPS-116}+\textit{INS}+\textit{PART}~(training) $\rightarrow$ PartImageNet~(evaluation).
As shown in Tab.~\ref{tab:ablation_np}, the object-part segmentation performance drops when the $N_p$ is too small~(1, 2) or too large~(4, 5, 6), . 

\begin{table}[ht]
\centering
\small
\renewcommand{\arraystretch}{1.2}
\begin{tabular}{c c c c c c c}
\toprule
$N_p$ & 1 & 2 & 3 & 4 & 5 & 6 \\
\midrule
Obj AP & 61.7 & 62.1 & 62.8 & 62.4 & 61.4 & 61.8 \\
Part AP & 15.4 & 15.8 & 16.4 & 16.0 & 16.1 & 15.9 \\
\bottomrule
\end{tabular}
\vspace{8pt}
\caption{Ablation Study on $N_p$.}
\label{tab:ablation_np}
\end{table}

\paragraph{Ablation on backbone finetuning.} 
\crv{We further conduct an ablation study to show the necessity of finetuning the visual backbone and pixel decoder during training. As we can see in the Tab.~\ref{tab:bk_pd_ablation}, finetuning the visual backbone and pixel decoder leads to improved performance especially in the part segmentation task. This effect is mainly due to the fact that the used visual backbone and pixel~\cite{cheng2021mask2former, maskdino} are pretrained only on object-level tasks, and the extracted dense features lack part-level understanding. Thus, finetuning them on the object-part-level tasks is beneficial. }

\begin{table}[t]
  \centering
  \scriptsize
  \begin{subtable}{\textwidth}
    \centering
    \begin{tabular}{l@{\hskip 10pt}ccc@{\hskip 10pt}ccc@{\hskip 10pt}ccc@{\hskip 10pt}ccc}
      \toprule
      \multirow{2}{*}{Method} 
      & \multicolumn{3}{c}{PPS-116} 
      & \multicolumn{3}{c}{+INS} 
      & \multicolumn{3}{c}{+INS+PART} 
      & \multicolumn{3}{c}{PartImageNet} \\
      & obj & part & AP & obj & part & AP & obj & part & AP & obj & part & AP \\
      \midrule
      Frozen Bk+Pd & 48.2 & 6.99 & 16.1 & 64.1 & 8.85 & 21.1 & 66.1 & 9.34 & 22.0 & 80.6 & 30.1 & 41.3 \\
      Frozen Bk    & 47.6 & 7.36 & 16.3 & 63.0 & 6.98 & 19.4 & 63.4 & 12.4 & 23.8 & 83.2 & 34.7 & 45.4 \\
      LangHOPS     & 49.1 & 8.62 & 17.6 & 61.8 & 13.6 & 24.3 & 62.7 & 17.0 & 27.1 & 85.5 & 47.9 & 55.8 \\
      \bottomrule
    \end{tabular}
    \vspace{2pt}
    \caption{(PPS-116 $\rightarrow$ PartImageNet.}
    \label{tab:bk_pd_ablation_a}
  \end{subtable}

  \vspace{8pt} 

  \begin{subtable}{\textwidth}
    \centering
    \begin{tabular}{l@{\hskip 10pt}ccc@{\hskip 10pt}ccc@{\hskip 10pt}ccc@{\hskip 10pt}ccc}
      \toprule
      \multirow{2}{*}{Method} 
      & \multicolumn{3}{c}{PartImageNet} 
      & \multicolumn{3}{c}{+INS} 
      & \multicolumn{3}{c}{+INS+PART} 
      & \multicolumn{3}{c}{PPS-116} \\
      & obj & part & AP & obj & part & AP & obj & part & AP & obj & part & AP \\
      \midrule
      Frozen Bk+Pd & 10.5 & 1.71 & 3.05 & 23.4 & 2.57 & 5.73 & 23.2 & 2.76 & 5.86 & 53.3 & 7.48 & 17.7 \\
      Frozen Bk    & 11.8 & 1.95 & 3.45 & 23.3 & 2.92 & 6.01 & 23.0 & 3.34 & 6.32 & 44.2 & 7.65 & 15.8 \\
      LangHOPS     & 11.3 & 2.17 & 3.47 & 21.9 & 3.82 & 6.55 & 23.8 & 4.16 & 7.13 & 56.4 & 15.3 & 21.4 \\
      \bottomrule
    \end{tabular}
    \vspace{2pt}
    \caption{PartImageNet $\rightarrow$ PPS-116.}
    \label{tab:bk_pd_ablation_b}
  \end{subtable}
  \caption{Ablations on frozen image backbones and pixel decoder in the cross-dataset settings. "BK" refers to the visual encoder and "Pd" refers to the pixel decoder in the Fig.~\ref{fig:our_framework}}
  \label{tab:bk_pd_ablation}
\end{table}

\subsection{Computation Cost} \label{subsec:suppl_computation_cost}
\crv{
We report the footprint of GPU hours, carbon cost, inference cost and model size of PartGLEE, PSALM and \OurMethod{}{}. 
The gpu hours and inference time are reported with Nvidia H200 GPU(s). The spec. power (700W) of H200 and world average carbon intensity of electricity (0.475 $kg\space CO_2 / kWh$ ) are used for calculating the footprint.
The Tab.~\ref{tab:computation_cost} shows that \OurMethod{}{} has the largest model size, mainly due to the usage of MLLM (Paligemma2-3B). 
PSALM† has the longest training time and carbon footprint since it trains the LLM instead of using LoRA, and needs to process all candidate category names, which leads to long input prompts to the LLM. 
\OurMethod{}{} achieves the best performance with reasonable training and inference cost compared to the baselines.}
\begin{table}[ht]
\centering
\small
\begin{tabular}{lcccc}
\toprule
\textbf{Method} & \textbf{Model Size} & \textbf{Training GPU Hours} & \textbf{Training Footprint (kg CO$_2$e)} & \textbf{Inference Time (ms)} \\
\midrule
PSALM$^{\dagger}$ & 1.5B & 92 & 30.6 & 628 \\
PartGLEE & 1B & 40 & 13.3 & 240 \\
LangHOPS & 4B & 72 & 23.9 & 396 \\
\bottomrule
\end{tabular}
\vspace{5pt}
\caption{Computation Cost of LangHOPS and the baselines. PPS116 + INS + PART $\rightarrow$ PartImageNet.}
\label{tab:computation_cost}
\end{table}

\clearpage

\newpage
\section*{NeurIPS Paper Checklist}

\begin{enumerate}

\item {\bf Claims}
    \item[] Question: Do the main claims made in the abstract and introduction accurately reflect the paper's contributions and scope?
    \item[] Answer: \answerYes{} 
    \item[] Justification: The abstract claims that the paper proposes a method which has key novelties in the model design and achieves state-of-the-part the performance. The introduction, method and experiment parts in the main paper clearly illustrate the model design, model's novelty and experiment implementation and demonstrate the contribution of the paper. 

\item {\bf Limitations}
    \item[] Question: Does the paper discuss the limitations of the work performed by the authors?
    \item[] Answer: \answerYes{} 
    \item[] Justification: It's in experiment section.

\item {\bf Theory assumptions and proofs}
    \item[] Question: For each theoretical result, does the paper provide the full set of assumptions and a complete (and correct) proof?
    \item[] Answer: \answerNA{} 
    \item[] Justification: The paper does not include theoretical results

    \item {\bf Experimental result reproducibility}
    \item[] Question: Does the paper fully disclose all the information needed to reproduce the main experimental results of the paper to the extent that it affects the main claims and/or conclusions of the paper (regardless of whether the code and data are provided or not)?
    \item[] Answer: \answerYes{} 
    \item[] Justification: The implementation details demonstrate the hyperparameters and training strategy. Supplementary material will provide further details due to the page limit. 

\item {\bf Open access to data and code}
    \item[] Question: Does the paper provide open access to the data and code, with sufficient instructions to faithfully reproduce the main experimental results, as described in supplemental material?
    \item[] Answer: \answerYes{} 
    \item[] Justification: The code will be released in Open Access, under some license.

\item {\bf Experimental setting/details}
    \item[] Question: Does the paper specify all the training and test details (e.g., data splits, hyperparameters, how they were chosen, type of optimizer, etc.) necessary to understand the results?
    \item[] Answer: \answerYes{} 
    \item[] Justification: It's in the method and experiment sections. 
    \item[] Guidelines:
    \begin{itemize}
        \item The answer NA means that the paper does not include experiments.
        \item The experimental setting should be presented in the core of the paper to a level of detail that is necessary to appreciate the results and make sense of them.
        \item The full details can be provided either with the code, in the appendix, or as supplemental material.
    \end{itemize}

\item {\bf Experiment statistical significance}
    \item[] Question: Does the paper report error bars suitably and correctly defined or other appropriate information about the statistical significance of the experiments?
    \item[] Answer: \answerYes{} 
    \item[] Justification: 

\item {\bf Experiments compute resources}
    \item[] Question: For each experiment, does the paper provide sufficient information on the computer resources (type of compute workers, memory, time of execution) needed to reproduce the experiments?
    \item[] Answer: \answerYes{} 
    \item[] Justification: We mention this in the implementation details.
    
\item {\bf Code of ethics}
    \item[] Question: Does the research conducted in the paper conform, in every respect, with the NeurIPS Code of Ethics \url{https://neurips.cc/public/EthicsGuidelines}?
    \item[] Answer: \answerYes{} 
    \item[] Justification: the paper conform, in every respect, with the NeurIPS Code of Ethics

\item {\bf Broader impacts}
    \item[] Question: Does the paper discuss both potential positive societal impacts and negative societal impacts of the work performed?
    \item[] Answer: \answerNA{} 
    \item[] Justification: 
    
\item {\bf Safeguards}
    \item[] Question: Does the paper describe safeguards that have been put in place for responsible release of data or models that have a high risk for misuse (e.g., pretrained language models, image generators, or scraped datasets)?
    \item[] Answer: \answerNA{} 
    \item[] Justification: no such risks

\item {\bf Licenses for existing assets}
    \item[] Question: Are the creators or original owners of assets (e.g., code, data, models), used in the paper, properly credited and are the license and terms of use explicitly mentioned and properly respected?
    \item[] Answer: \answerYes{} 
    \item[] Justification: The models and data used in the paper are properly credited and the license and terms of use are properly respected.

\item {\bf New assets}
    \item[] Question: Are new assets introduced in the paper well documented and is the documentation provided alongside the assets?
    \item[] Answer: \answerYes{} 
    \item[] Justification: They will be provided upon acceptance of the paper.

\item {\bf Crowdsourcing and research with human subjects}
    \item[] Question: For crowdsourcing experiments and research with human subjects, does the paper include the full text of instructions given to participants and screenshots, if applicable, as well as details about compensation (if any)? 
    \item[] Answer: \answerNA{} 
    \item[] Justification: The paper does not involve crowdsourcing nor research with human subjects.

\item {\bf Institutional review board (IRB) approvals or equivalent for research with human subjects}
    \item[] Question: Does the paper describe potential risks incurred by study participants, whether such risks were disclosed to the subjects, and whether Institutional Review Board (IRB) approvals (or an equivalent approval/review based on the requirements of your country or institution) were obtained?
    \item[] Answer: \answerNA{} 
    \item[] Justification: The paper does not involve crowdsourcing nor research with human subjects.

\item {\bf Declaration of LLM usage}
    \item[] Question: Does the paper describe the usage of LLMs if it is an important, original, or non-standard component of the core methods in this research? Note that if the LLM is used only for writing, editing, or formatting purposes and does not impact the core methodology, scientific rigorousness, or originality of the research, declaration is not required.
    \item[] Answer: \answerYes{} 
    \item[] Justification: The proposed model leverages a open-source MLLM as the component of the framework.  

\end{enumerate}

\end{document}